\documentclass{article}

\usepackage[preprint]{neurips_2024}

\usepackage[utf8]{inputenc} % allow utf-8 input
\usepackage[T1]{fontenc}    % use 8-bit T1 fonts
\usepackage[hidelinks]{hyperref}       % hyperlinks
\usepackage{url}            % simple URL typesetting
\usepackage{booktabs}       % professional-quality tables
\usepackage{amsfonts}       % blackboard math symbols
\usepackage{nicefrac}       % compact symbols for 1/2, etc.
\usepackage{microtype}      % microtypography
\usepackage{xcolor}         % colors
\usepackage{graphicx}
\usepackage{algorithm}
\usepackage{algpseudocode}
\usepackage{amsmath}
\usepackage{multirow}
\usepackage{xspace}
\usepackage[autolanguage,np]{numprint}

\title{Efficient Minimum Bayes Risk Decoding using Low-Rank Matrix Completion Algorithms}

\newcommand{\bleurt}{\textsc{Bleurt}\xspace}
\newcommand{\metricx}{MetricX\xspace}

\newcommand{\comet}{\textsc{Comet22}\xspace}
\newcommand{\bleu}{\textsc{Bleu}\xspace}
\newcommand{\chrf}{ChrF\xspace}

\newcommand{\lpfromto}{$\leftrightarrow$}
\DeclareMathOperator*{\argmax}{argmax}

\newcommand{\ombr}{FMBR\xspace}
\newcommand{\nxk}{$N\!\!\times\!K$}
\newcommand{\sxs}{$S\!\!\times\!\!S$}

\author{%
  Firas Trabelsi \\
  Google\\
  \texttt{firast@google.com} \\
  \And
   David Vilar \\
   Google \\
   \texttt{vilar@google.com} \\
   \And
   Mara Finkelstein \\
   Google \\
  \texttt{marafin@google.com} \\
   \And
   Markus Freitag \\
   Google \\
   \texttt{freitag@google.com} \\
}

\begin{document}

\maketitle

\begin{abstract}
Minimum Bayes Risk (MBR) decoding is a powerful decoding strategy widely used for text generation tasks, but its quadratic computational complexity limits its practical application.
This paper presents a novel approach for approximating MBR decoding using matrix completion techniques, focusing on the task of machine translation.
We formulate MBR decoding as a matrix completion problem, where the utility metric scores between candidate hypotheses and pseudo-reference translations form a low-rank matrix.
First, we empirically show that the scores matrices indeed have a low-rank structure.
Then, we exploit this by only computing a random subset of the scores and efficiently recover the missing entries in the matrix by applying the Alternating Least Squares (ALS) algorithm, thereby enabling a fast approximation of the MBR decoding process.
Our experimental results on machine translation tasks demonstrate that the proposed method requires 1/16 utility metric computations compared to vanilla MBR decoding while achieving equal translation quality measured by \comet on the WMT22 dataset (en \lpfromto de, en \lpfromto ru).
We also benchmark our method against other approximation methods and we show gains in quality when comparing to them.
\end{abstract}

\section{Introduction}

The generation process in most conditional natural language processing tasks is usually guided by the maximum-a-posteriori (MAP) rule: given an input $x$, generate the output $\hat{y}$ that maximizes the posterior probability distribution: $\hat{y} = \argmax_y p(y|x)$.
%An exact maximization usually involves searching through an exponentially large search space, e.g.\ the space of all possible sentences in a given language.
%As an exhaustive search is computationally intractable, typically beam search or greedy decoding are used to approximate the search for the best hypothesis.
%
It can be shown that MAP decoding is optimal under a 0-1 loss criterion.
However for more nuanced tasks, where different outputs can be considered correct, and the quality of the output is not just a binary decision between ``right'' and ``wrong'', MAP decoding is no longer optimal.
Neural Machine Translation~(NMT) is a prominent example of these types of tasks.
For NMT, a system is trained to generate a sentence in a target language given a source sentence in another language.
For a given sentence, there exists a variety of possible translations, each of which has a different quality level.
\citet{eikema-aziz-2020-map} demonstrated that MAP decoding methods are suboptimal for NMT, and showed that other generation strategies may be preferred.
Furthermore, NMT models often assign human translations lower probabilities than their own beam search outputs, due to model calibration issues~\citep{ott2018analyzing,freitag2020bleu}.

As an alternative, \citet{eikema-aziz-2020-map, eikema2022samplingbased} applied MBR decoding for NMT models.
MBR decoding follows a two-step approach, where a model is used to generate a list of candidate translations and a list of pseudo-references (which may be the same as the list of candidates).
The candidates are then scored with a performance metric using the pseudo-references as an approximation of the true references, and the candidate with the maximum expected quality (or equivalently minimum risk) is then selected.
In contrast to MAP decoding, MBR decoding is not designed to generate the translation with the highest estimated model probability; instead it aims to directly optimize a utility function.
Subsequent research conducted by \citet{freitag2022high} showed that MBR decoding with \emph{neural} utility metrics leads to significant improvements over beam search decoding.
However, MBR is computationally expensive, with a time complexity of $O(N^2)$ for a candidate list containing $N$ samples and $N$ pseudo-references (usually the two lists coincide).
According to \cite{freitag2022high}, ideally $N$ ranges between $100$ and $1\,000$, which involves thousands to millions of utility function computations.
Note than when using neural metrics, each of the $O(N^2)$ ``computation steps'' is itself expensive, requiring a forward pass through a large neural network.
%Thus, reducing the number of computations needed for the process has an important impact on the overall efficiency of the decoding algorithm.

In this work, we propose to reduce the number of metric computations by scoring only a subset of candidate--pseudo-reference pairs.
We then proceed to use a matrix completion algorithm (ALS in our case) to estimate the remaining utility scores.
For such completion algorithms to work, the full matrix has to fulfill some conditions, specifically to be a low-rank matrix.
We empirically show that this is indeed the case for the utility matrices for MBR decoding.
Intuitively, this makes sense: akin to recommendation systems, where similar users are expected to have similar book or movie preferences, similar pseudo-references are expected to have similar ``translation preferences''.
Experimental results show the effectiveness of our method, where the performance of the full MBR algorithm can be matched with only a fraction of the computational cost.
Compared to other approaches that also seek to reduce the number of computations, our method does not compromise translation quality, as confirmed by human evaluation.

Our scientific contributions are as follows:
\begin{enumerate}
  \item We empirically show that the utility matrices for MBR decoding are low-rank.
  \item We apply ALS to a subset of scores to approximate the full MBR matrix.
  \item We show that using our method we can reduce the number of computations by a factor of 16, while maintaining the same translation quality level.
\end{enumerate}

\section{Related Work}

% Machine Translation~(MT) metrics usually are based the comparison between a candidate translation with a reference translation for the given source sentence, usually human-generated.
% A multitude of reference-based metrics are available to evaluate the quality of translated content.
% Some metrics rely on lexical overlap, such as \bleu \citep{papineni-etal-2002-bleu}, \meteor \citep{banerjee-lavie-2005-meteor} or \chrf \citep{popovic-2015-chrf}.
% The WMT metrics task~\citep{freitag-etal-2022-results} demonstrated that the new generation of metrics -- neural fine-tuned metrics like \bleurt \citep{sellam2020bleurt} and \comet \citep{rei-etal-2020-comet} -- have significantly higher correlation with human judgement than traditional word overlap metrics.
% Consequently, we focus on neural fine-tuned metrics in this study.
% A prominent example of a reference-based neural metric is \bleurt \citep{sellam2020bleurt}, and its extension to \metricx, which was the winning entry in the WMT22 metrics task \citep{freitag-etal-2022-results}.

While MT research has traditionally relied on MAP decoding or generating $k$-best lists through beam search for MBR decoding, \citet{eikema-aziz-2020-map} proposed an approximation of MBR decoding via unbiased sampling.
Their method aims to address the limitations of MAP decoding \citep{eikema-aziz-2020-map, muller-sennrich-2021-understanding, eikema2022samplingbased} by demonstrating that samples drawn from the NMT model align more faithfully with training data statistics when compared to beam search.
\citet{freitag2022high} showed that using neural metrics results in significant improvements in translation quality.
\citet{freitag2023epsilon} reported that the choice of sampling approach is important, and epsilon sampling \citep{hewitt-etal-2022-truncation} is ideal for MBR decoding and reranking.

While the improvements in translation quality afforded by MBR are widely acknowledged, its high computational cost limits its application in practice.
Different approaches have been proposed to speed up MBR computation.
\citet{eikema2022samplingbased} propose to decouple the candidate and pseudo-reference lists to allow for different sizes, and propose a coarse-to-fine refinement of the hypothesis space.
%We take their approach as a baseline with which to compare our method.
\citet{cheng2023faster} speed up MBR decoding by gradually increasing the number of samples used to estimate the utility, while additionally pruning the hypothesis space.
\citet{jinnai2024hyperparameterfree} formulate MBR as a medoid identification problem, and apply approximate algorithms developed on this problem.
\cite{vamvas2024lineartime} aggregate the set of pseudo-references, allowing for just one utility computation per candidate.
This greatly accelerates the decoding process, but the utility metric needs to fulfill certain conditions to be applicable.
\cite{finkelstein2024mbr} use MBR decoding in a knowledge-distillation framework to simulate MBR decoding with single-pass search.
\cite{tomani2024qualityaware} train quality-aware translation models in order to reduce the size of the candidate list.
Similar in spirit to MBR decoding, QE-rescoring approaches~\citep{fernandes-etal-2022-quality} also directly optimize a utility function, with linear-time cost.

Low-Rank Matrix completion is an active area of research and multiple algorithms have been developed to perform it.
\citep{8759045} is an extensive survey for such methods.
Some of the popular algorithms are: Singular Value Thresholding ~\citep{cai2008singular}, Bayesian Probabilistic Matrix Factorization ~\citep{7830213}, Maximum Margin Matrix Factorization\citep{NIPS2004_e0688d13} and Alternating Least Squares \citep{Zachariah}, which is the one we chose for this work.
To the best of our knowledge, this work is the first one to apply matrix completion algorithms for completing a partial MBR score matrix.

\section{Preliminaries}

We are given a NMT model $P_{\Theta}(y|x)$ which serves to estimate the probability of a hypothesis segment~$y$, given a source segment~$x$, with $\Theta$ being the learned parameters of the neural network.
MAP decoding  involves searching for the most probable translation under $P_{\Theta}(y|x)$.
However, determining the hypothesis with the maximum probability is computationally intractable due to the expansive and combinatorially complex search space.
Consequently, approximations like beam search \citep{graves2012sequence, sutskever2014sequence} are often employed.

If we want to generate diverse hypotheses, e.g.\ in generative tasks where creativity is desired instead of selecting the candidate with the highest probability (or an approximation thereof), we sample the output sentence following the probability distribution defined by the model.
For NMT, this approach is used for generating a list of candidate translations.
Specifically, epsilon sampling, as outlined by \citet{hewitt-etal-2022-truncation}, has emerged as the leading sampling technique for MBR. It was shown by \citet{freitag2023epsilon} to outperform other methods such as ancestral, top-$k$ or nucleus sampling \citep{holtzman2019curious}.
Epsilon sampling prunes away any token with a probability lower than a threshold $\varepsilon$, thereby guaranteeing that each token within a sample is allocated a fair probability mass.
%The likelihood of selecting token $y^{(\tau)}$ in the sampling process at time $\tau$ is governed by
%\begin{equation}
%    P'_{\Theta, \varepsilon} (y^{(\tau)}|x, y^{(1:\tau-1)}) \sim \begin{cases}
%    p_\tau^{}{}^{\frac{1}{T}} & \text {if } p_\tau \geq \varepsilon \\
%    0 & \text{otherwise}
%    \end{cases},
%\end{equation}
%with
%\begin{equation*}
%p_\tau = P_{\Theta}(y^{(\tau)}|x, y^{(1:\tau-1)}).
%\end{equation*}
%$T$ denotes the sampling temperature. %, determining whether the sampling distribution exhibits a flat or peaked shape.
%Epsilon sampling proves to be a highly effective strategy for the selective removal of unreliable, low-probability tokens.

\subsection{Minimum Bayes Risk Decoding}
\label{sec:mbr}

In MBR decoding \citep{statistics1977basic,Berger_decision_theory_1985}, given a set of candidate hypotheses \(\mathcal{H}\), the goal is to select the optimal hypothesis based on its expected utility, measured by a function $u$, with respect to the distribution over human references within the space of all references $\mathcal{Y}$.
%This can be expressed mathematically as:
%\begin{eqnarray}
%\label{eq:true_expected_utility}
%h^{\rm best}
%& = &  \argmax_{y\in \mathcal{H}} \sum_{r\in \mathcal{Y}} u(y, r) P_{\rm human}(r|x),
%\end{eqnarray}
%where $u(y, r)$ is a utility metric that is being used to gauge the quality of a candidate translation $y$ with respect to a reference translation $r$.

Since the true distribution remains unknown, we resort to sampling from the model instead, which relies on the assumption that the model provides a reliable approximation for the true underlying distribution over human translations.
Furthermore, the integration over the vast space of all possible references $\mathcal{Y}$ is computationally intractable.
Therefore, MBR adopts a finite sample estimate by sampling a set of pseudo-references $\mathcal{R}$ from \(P_{\theta}(\cdot|x)\). This approximation can be expressed as:
\begin{equation}
\label{eq:approx_model_expected_utility}
h^{\rm MBR} = \argmax_{h \in {\cal H}} \frac{1}{|\mathcal{R}|} \sum_{r\in \cal \mathcal{R}} u(h, r) .
\end{equation}
Usual practice is to set $\mathcal{H} = \mathcal{R}$, i.e.\ the same set of model hypotheses serves both as the candidate list~$\mathcal{H}$ as well as the pseudo-reference list~$\mathcal{R}$.
The computational time complexity of MBR decoding is \(O(N^2)\) with $N$ the size of the candidate list.

Note that this quadratic expression refers to \emph{each sentence} to translate, i.e. for a corpus of size $S$, the total cost will be $O(S \cdot N^2)$.
Also there is a hidden (multiplicative) constant, namely the cost of the computation of the utility function.
For surface level metrics (e.g.\ \bleu, \chrf), this cost is negligible, but for neural metrics it involves computing the forward pass of a large neural network.
%thus every reduction in metric computations has an important effect on the total running cost.
Therefore, any reduction in the number of metric computations has an important effect on the total running cost.

\subsection{Low-Rank Matrix Completion}

Low-Rank Matrix Completion is a fundamental problem in machine learning and data analysis with popular application such as \textit{Collaborative Filtering} \citep{10.1145/1102351.1102441} and \textit{Image Denoising} \citep{candes2008exact}.
The goal of matrix completion is to estimate the missing entries of a partially observed matrix, under the assumption that the underlying matrix is low-rank.
This assumption implies that the matrix can be well-approximated by a product of two smaller matrices, capturing the latent factors that explain the observed data.
\citet{candes2008exact} proved that perfect approximations can be achieved if the number of observed entries is larger than $C N^{1.2} r \log(N)$ for some positive numerical constant $C$, for most $N \times{} N$ matrices of rank~$r$ with very high probability. 

One simple and efficient algorithm is Alternating Least Squares (ALS) \citep{Zachariah}.
To recover any matrix $M$, the algorithm approximates it by two smaller matrices $M \approx X^TY$ and then minimizes the following equation given the observed entries.
\begin{equation}
\min_{X,Y} \sum_{m_{ij} \text{ observed}} (m_{ij} - x_i^T y_j)^2 + \lambda \left( \sum_i ||x_i||^2 + \sum_j ||y_j||^2 \right)
\end{equation}
The algorithm achieves this by alternatively solving for $X$ and $Y$ as shown in Algorithm~\ref{alg:als} 
The algorithm has three hyperparameters: $\lambda$ a regularization term, $r$ the second dimension of the smaller matrices and $n$ the number of alternating steps performed. The main motivation for picking this algorithm in our approach is its simple implementation.
\begin{algorithm}
\caption{ALS for Matrix Completion}
\label{alg:als}
\begin{algorithmic}[1]

\Require $\lambda$, $r$ and $n$
\State Initialize $X$, $Y$ with shapes $N\times r $ and $r\times N$
\Repeat
    \For{$i = 1 \dots n$}
        \State $x_i = \left( \sum_{m_{ij} \in m_{i*}} y_j y_j^\top + \lambda I_k \right)^{-1} \sum_{m_{ij} \in m_{i*}} m_{ij} y_r$ \hfill
    \EndFor
    \For{$j = 1 \dots n$}
        \State $y_j = \left( \sum_{m_{ij} \in m_{*j}} x_r x_r^\top + \lambda I_k \right)^{-1} \sum_{m_{ij} \in m_{*j}} m_{ij} x_i$ \hfill
    \EndFor
\Until{convergence}
\end{algorithmic}
\end{algorithm}

\section{MBR Matrix}
\label{gen_inst}

\subsection{Definition of MBR matrix}

Given a source sentence, we use an NMT model to generate a set $\mathcal{H}$ of hypotheses such that $|\mathcal{H}| = N$.
As explained in the preliminaries section, the MBR method uses two different sets of hypotheses and pseudo-references, but in practice we use the same set of samples for both $\mathcal{H}$ and $\mathcal{R}$.
The pairwise scores for all hypotheses in $\mathcal{H}$ gives an $N\times N$ matrix $M$ such that $M[i,j] = U(h_i, h_j)$ for all $(h_i,h_j) \in \mathcal{H} \times \mathcal{H}$ and a utility metric $U$ that computes some similarity between two hypotheses.
With this matrix formulation, MBR decoding reduces to picking the row with the highest average  (since each row maps to one sample in the hypotheses list).

\subsection{MBR matrices are low rank}

Intuitively, the values within the MBR matrix are highly correlated, since by definition each value $M[i,j]$ is computing a similarity score between two hypotheses given a utility metric.
This is a key assumption that our work is built on since low-rank matrices have theoretical bounds on the number of entries required to recover the full matrix \citep{candes2008exact}.

\begin{table}
\label{table:svd}
\caption{Summary of the first three singular values of MBR matrices for the MetricX and chrF utility functions, with two different sizes and four different language pairs}
  \centering
  \newlength{\colskip}
  \setlength{\colskip}{7mm}
  \setlength{\tabcolsep}{4pt}
  \begin{tabular}{cccc@{\hskip \colskip}ccc@{\hskip \colskip}ccc@{\hskip \colskip}ccc}
    \toprule
    & \multicolumn{6}{c}{64x64}  & \multicolumn{6}{c}{128x128} \\
    \cmidrule(r@{\colskip}){2-7} \cmidrule(l){8-13}
    & \multicolumn{3}{c@{\hskip \colskip}}{MetricX} & \multicolumn{3}{c@{\hskip \colskip}}{chrF}   & \multicolumn{3}{c@{\hskip \colskip}}{MetricX} & \multicolumn{3}{c}{chrF} \\
    \cmidrule(lr{\colskip}){2-4} \cmidrule(r{\colskip}){5-7} \cmidrule(lr{\colskip}){8-10} \cmidrule(r){11-13}
    LP     & $\sigma_1$  & $\sigma_2$ & $\sigma_3$  & $\sigma_1$  & $\sigma_2$ & $\sigma_3$ & $\sigma_1$  & $\sigma_2$ & $\sigma_3$ & $\sigma_1$  & $\sigma_2$ & $\sigma_3$\\
    \midrule
    English$\to$German & 45.7 & 2.1 & 1.0 & 39.9 & 2.4  & 1.4  & 91.6 & 3.8 & 1.7 & 76.2 & 4.2 & 2.1 \\
    German$\to$English & 47.4 & 2.1 & 1.4 & 47.5 & 1.5  & 1.4  & 94.6 & 3.7 & 2.1 & 93.7 & 3.0 & 2.0 \\
    English$\to$Russian & 47.7 & 2.0 & 1.1 & 36.1 & 1.9  & 1.3  & 95.4 & 3.6 & 1.7 & 76.8 & 3.4 & 2.1 \\
    Russian$\to$English & 46.1 & 2.2 & 1.0 & 40.5 & 2.8  & 1.3  & 92.0 & 4.1 & 1.7 & 81.8 & 5.7 & 2.1 \\
    \midrule
    Average   & 46.7 & 2.1 & 1.1 & 41.0 & 2.15 & 1.35 & 93.4 & 3.8 & 1.8 & 82.1 & 4.1 & 2.1 \\
    \bottomrule
  \end{tabular}

\end{table}

\begin{figure}
  \centering
  \includegraphics[width=0.8\columnwidth]{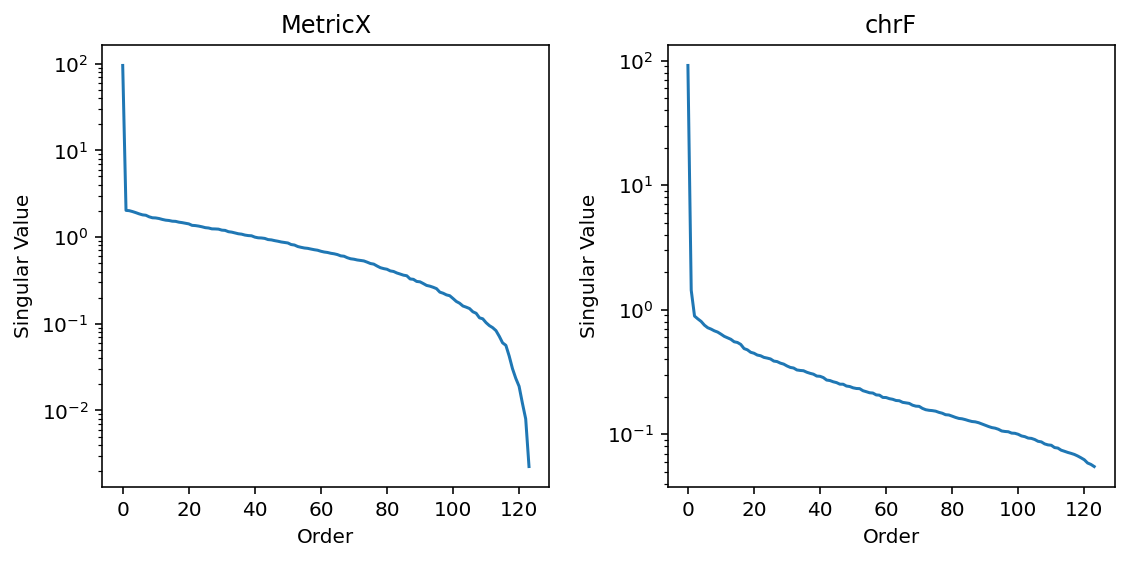}
  \label{fig:svd}
  \caption{Plot the singular values of an example 124x124 MBR matrix using logscale. We observe a sharp drop after the first singular value for the two utility metrics indicating that the matrix is  rank-1.}
\end{figure}

We verify this assumption empirically. We generated 1024 samples for each example in the WMT 2022 en \lpfromto de and en \lpfromto ru datasets.
We then generated the $N\times N$ matrices for different values of $N \in \{64,128\}$ by only considering a random subset of the samples using two different utility metrics: \metricx \citep{freitag-etal-2022-results} and \chrf \citep{popovic-2015-chrf}.
We then perform singular value decomposition and look at the distribution of the singular values, shown in Figure~\ref{fig:svd} and Table~\ref{table:svd}.
We observe that across utility metrics and matrix dimensions, \textbf{$\sigma_1 \gg \sigma_2$}.
On average across datasets we have $\sigma_2 / \sigma_1 < 0.05 $, this means that most of the information within the matrix can be captured by a single dominant direction or component and thus can be approximated by a rank-1 matrix.

\section{The PMBR Method}
\label{sec:pmbr-method}
We propose an approximation method for MBR decoding that leverages the low-rank structure of the MBR matrix.
The procedure is shown in Algorithm \ref{alg:pmbr}.
Given an NMT model, we start by generating a set of hypotheses $\mathcal{H}$ similar to the vanilla MBR method.
Then, instead of computing all the pairwise scores in the utility matrix, we only compute a random subset of the scores that we denote with $\Omega$.
The size of $\Omega$ depends on the \emph{computation budget} available.
We define the budget as the ratio of computations performed with respect to the total amount of computations to compute the full matrix.
Thus for any given budget $1/r$, we end up with $N^2 / r$ entries observed in the matrix.
The next step is to apply ALS on $M[\Omega]$ as described in Algorithm~\ref{alg:als}, where $M[\Omega]$ denotes the matrix of size $N \times N$ where only the entries of $\Omega$ are non-null.
Finally, with all the pairwise scores recovered, we perform vanilla MBR decoding.
We call this procedure PMBR for \emph{Probabilistic MBR} decoding.

\begin{algorithm}
\caption{PMBR: MBR Approximation using ALS}
\label{alg:pmbr}
\begin{algorithmic}[1]

\Require List of hypotheses $H$, reduction ratio $r \in (0,1)$

\State $N \gets |H|$
\State $S \gets \lceil |N|^2 \cdot r \rceil$ \Comment{Number of utility computations}
\State $\Omega \gets$ Sample $S$ coordinate pairs $(i, j)$ from $N \times N$
\State $M \gets \mathbf{0}^{|N| \times |N|}$ \Comment{Initialize empty matrix}
\For{$(i, j) \in \Omega$}
    \State $M_{ij} \gets U(i, j)$
\EndFor
\State $M \gets \text{ALS}(M, \text{hyperparameters})$

\State \Return \text{vanilla\_MBR(M)}
\end{algorithmic}
\end{algorithm}

\paragraph{Time Complexity}
The time complexity of this algorithm is dominated by the utility metrics computations.
The utility metrics are deep neural networks that require $O(\mathit{millions})$ matrix multiplications while ALS requires $O(\mathit{hundreds})$ matrix multiplications.
For reference, 30 steps of the ALS algorithm with $r=10$ running on a CPU takes on average 0.2 seconds to run while the MetricX inference takes 3.4 seconds on a TPUv4 platform.
Thus, the savings in run time achieved by our approximation is close to proportional to the savings in number of utility computations.
Note that this analysis focuses only on the second stage of MBR decoding, i.e.\ we do not take the cost of generating the hypotheses into account.

\section{Experimental Setup}
\label{sec:experiments}

\subsection{Metrics}
We use \metricx \citep{juraska-etal-2023-metricx} as the utility function for all variants of MBR decoding as it has been shown that neural fine-tuned metrics outperform word-overlap metrics like \bleu~\citep{papineni-etal-2002-bleu} and \chrf~\citep{popovic-2015-chrf} for MBR decoding \citep{freitag2022high}. \metricx is an extension of \bleurt~\citep{sellam2020bleurt}, showing higher correlation with human judgment \citep{freitag-EtAl:2023:WMT} and has been designed to also work on multi-sentence segments \citep{deutsch-etal-2023-training} and not only sentences in isolation.
In addition, we report \comet \citep{rei-etal-2020-comet,rei-etal-2022-comet} scores as there is a risk of overfitting \citep{amrhein-sennrich-2022-identifying} on \metricx.
In addition, for one selected experiment we conducted expert-based human evaluations using MQM \citep{freitag-etal-2021-experts}, a human evaluation scheme centered on marking errors present in the translations.
%~ To show that our method generalizes to different settings, we change two variables: the budget $1/r$ and the size of the hypothesis list $N$.
We report results by varying the budget available to the MBR methods.
For each budget, we randomly sample from the full MBR matrix, and report the average results of 1000 trials.

\subsection{Datasets and Model}

We run experiments using the WMT 2022 test sets for English\lpfromto German (en\lpfromto de) and English\lpfromto Russian (en\lpfromto ru). The official WMT test sets \citep{kocmi-etal-2022-findings} are split into sentences but come with document information. We constructed multi-sentence (paragraph) level test sets with the following method:
For each document, we concatenate sentences together as long as we do not exceed 500 SPM tokens (given the \metricx SPM model). We respect sentence boundaries and do not truncate sentences.
Note: In WMT22, we have 4 different domains. Some domains have no document context and the segments remain single sentence segments also in the multi-sentence test sets.
Test data statistics can be seen in Appendix~\ref{table:testdata_stats}.
We use PaLM8B \citep{chowdhery2022palm} as translation model and sample 1024 examples for each sentence using epsilon sampling with $\epsilon = 0.02$ \citep{freitag2023epsilon} and using 3-shot prompting with examples taken from the FLORES corpus \citep{guzman-etal-2019-flores}

\subsection{Decoding Methods}
\label{sec:decoding-methods}
We compare our approximation PMBR against three other decoding methods.
To enable a fair comparison, we adapt each method so that the number of utility function computations is the same for each method.
(Recall that for a given budget $1/r$, we only observe $N^2/r$ entries in the matrix when performing PMBR.)
We compare PMBR with the following methods:
\begin{itemize}
  \item \ombr: This is the full MBR method. This is the only method that is not affected by the budget i.e the full matrix is observed.
    \item \nxk: This method was proposed by \citet{eikema2022samplingbased} and works by shrinking the pseudo-references list size.
      For a budget $1/r$ the MBR matrix gets reduced to an $N\times{}K$  matrix with $K=N/r$. The $K$ pseudo-references are randomly sampled.
    \item \sxs: This method corresponds to \ombr, but reduces the total size of the utility matrix to a size of $S \cdot S$, where the total number of entries corresponds to the available budget, i.e.\ $S=\sqrt{N^2/r}$. The $S$ examples are randomly sampled.
\end{itemize}

\subsection{Hyperparameter Tuning}
The ALS algorithm has three hyperparameters $\lambda$, $n$ and $r$ as described in Algorithm \ref{alg:als}.
We perform a grid search to optimize these hyperparameters, setting our loss function to be the accuracy with respect to the vanilla MBR method.
Concretely, for each example sentence we rank all samples by running the vanilla MBR.
Let us denote with $h_\text{MBR}$ the hypothesis selected by the full MBR method, and let $\operatorname{pos}_\text{PMBR}(h_\text{MBR})$ the rank the position of $h_\text{MBR}$ after ordering the hypotheses according to the scores predicted by PMBR.
The loss function is then just the sum of $\operatorname{pos}_\text{PMBR}(h_\text{MBR})$ for all the hypotheses in a subset of the data.
%Let us denote this ranking by a list~$G$.
%Then if PMBR picks sample $i$, the loss is equal to $j$ such that $G[j] = i$.
%\todo{I (vilar@) don't fully understand this. What is $j$? Is this a loss (minimization) or an accuracy (maximization)?}
We minimize this loss per language pair on 10 examples that we hold from the data generated with the WMT 2022 datasets with this search space $\{\lambda \in \{0.1, 0.15, 0.2\}\} \times \{r \in \{5, 6, \ldots 14, 15\}\} \times \{n \in \{10, 11, \ldots, 29, 30\}\}$

\begin{figure}
  \centering
  \label{fig:deen_scores}
  \includegraphics[width=0.8\columnwidth]{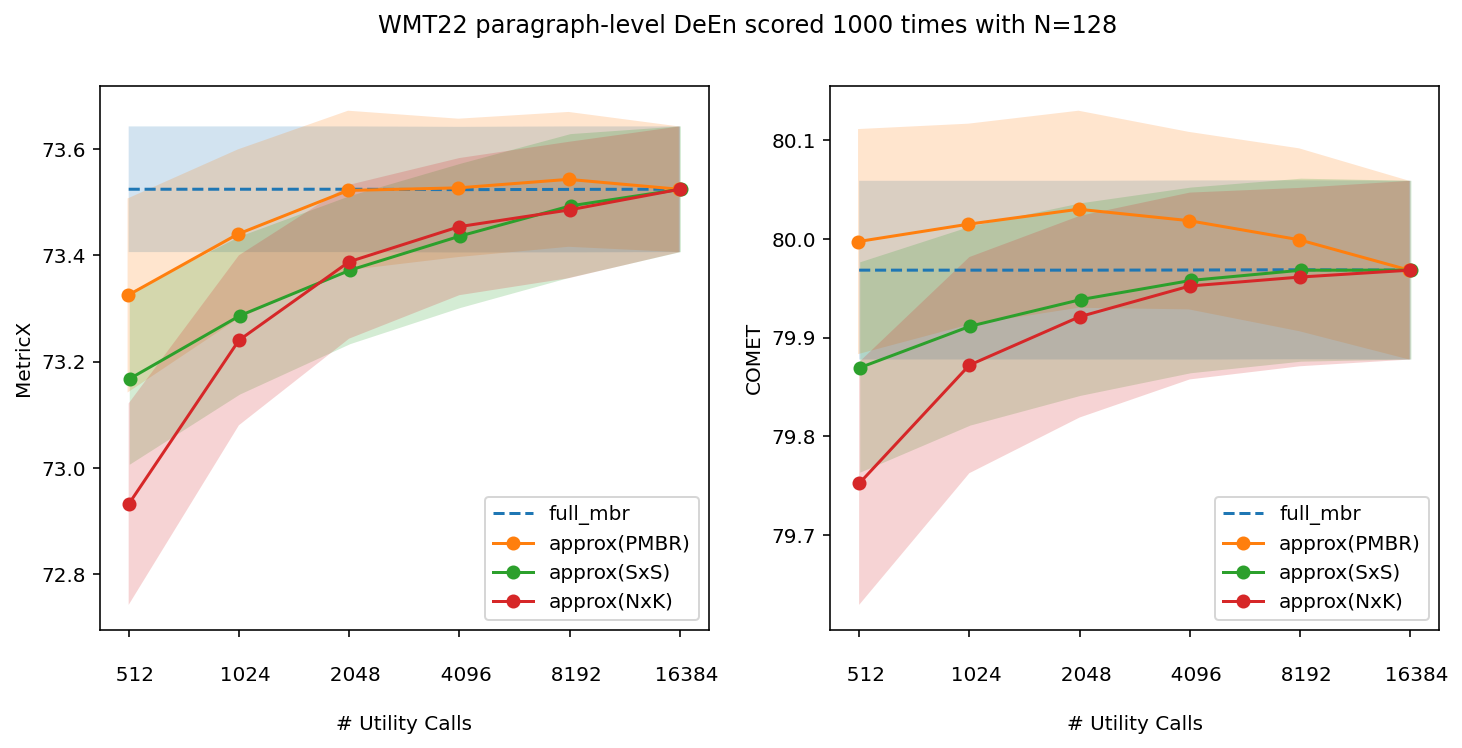}
  
  \begin{tabular}{llcccccc}
    \toprule
    & budget            & $1/32$         & $1/16$         & $1/8$          & $1/4$          & $1/2$          & $1/1$            \\
    & \# utility calls  & $512$          & $1024$         & $2048$         & $4096$         & $8192$         & $16384$          \\
    \midrule
    \multirow{3}{*}{\begin{minipage}{1cm}\centering Matrix size\end{minipage}}
    & PMBR ($|\Omega|$) & $512$          & $1\,024$       & $2\,048$       & $4\,096$       & $8\,192$       & $16\,384$        \\
    & \nxk              & $128\times 4$  & $128\times 8$  & $128\times 16$ & $128\times 32$ & $128\times 64$ & $128\times 128$  \\
    & \sxs              & $22 \times 22$ & $32 \times 32$ & $45 \times 45$ & $64 \times 64$ & $90 \times 90$ & $128 \times 128$ \\
    \bottomrule
  \end{tabular}

  \caption{We scored WMT22 DeEn dataset 1000 times for each budget available. Each scoring picks without replacement 128 samples from the 1024 samples available for each sentence. The highlighted area shows the standard deviation of the scores.}
\end{figure}

\begin{table}
  \caption{Results on the four translation directions on the WMT22 data.
    Each number (except for \ombr) is the average of 1\,000 runs with different random values taken from the full MBR matrix.
    $N$ is set to 128, and the budget is allocated according to the description in Section~\ref{sec:decoding-methods}.
    `C' denotes \comet scores and `X' \metricx scores.
  }
  \label{table:all_lps}

  \small
  \centering
  \setlength{\colskip}{7mm}
  \setlength{\tabcolsep}{4pt}
  \nprounddigits{3}
  \begin{tabular}{ll
      %n{1}{3}n{1}{3}@{\hskip \colskip}
      %n{1}{3}n{1}{3}@{\hskip \colskip}
      %n{1}{3}n{1}{3}@{\hskip \colskip}
      %n{1}{3}n{1}{3}@{\hskip \colskip}
      %n{1}{3}n{1}{3}
      ll@{\hskip \colskip}
      ll@{\hskip \colskip}
      ll@{\hskip \colskip}
      ll@{\hskip \colskip}
      ll
    }
\toprule
                                                & budget & \multicolumn{2}{c@{\hskip \colskip}}{1/32} & \multicolumn{2}{c@{\hskip \colskip}}{1/16} & \multicolumn{2}{c@{\hskip \colskip}}{1/8} & \multicolumn{2}{c@{\hskip \colskip}}{1/4} & \multicolumn{2}{c}{1/2} \\

  \cmidrule(lr{\colskip}){3-4}
  \cmidrule(r{\colskip}){5-6}
  \cmidrule(r{\colskip}){7-8}
  \cmidrule(r{\colskip}){9-10}
  \cmidrule(r){11-12}

                                                &        & \multicolumn{1}{c}{C}                   & \multicolumn{1}{c@{\hskip \colskip}}{X}                 & \multicolumn{1}{c}{C}                  & \multicolumn{1}{c@{\hskip \colskip}}{X}                & \multicolumn{1}{c}{C}                   & \multicolumn{1}{c@{\hskip \colskip}}{X}        & \multicolumn{1}{c}{C}          & \multicolumn{1}{c@{\hskip \colskip}}{X}        & \multicolumn{1}{c}{C}          & \multicolumn{1}{c}{X}        \\
\midrule
\multirow{4}{*}{\rotatebox[origin=c]{90}{en$\to$de}} & \ombr & 83.52                   & 77.01                   & 83.52                  & 77.01                  & 83.52                   & 77.02          & 83.52          & 77.02          & 83.52          & 77.01          \\
\cmidrule{2-12}
                                                & PMBR   & \textbf{83.53}          & 75.94                   & \textbf{83.63}         & \textbf{76.50}         & \textbf{83.63}          & \textbf{76.81} & \textbf{83.60} & \textbf{76.96} & \textbf{83.56} & \textbf{77.01} \\
                                                & \nxk    & 82.18                   & 74.96                   & 82.90                  & 75.99                  & 83.28                   & 76.59          & 83.45          & 76.84          & 83.48          & 76.93          \\
                                                & \sxs    & 83.39                   & \textbf{76.12}          & 83.52                  & 76.43                  & 83.57                   & 76.63          & 83.59          & 76.79          & 83.57          & 76.91          \\
\midrule

\multirow{4}{*}{\rotatebox[origin=c]{90}{de$\to$en}} & \ombr & 79.97                   & 73.52                   & 79.97                  & 73.52                  & 79.97                   & 73.52          & 79.97          & 73.52          & 79.97          & 73.52          \\
 \cmidrule{2-12}
                                                & PMBR   & \textbf{80.00}          & \textbf{73.32}          & \textbf{80.01}         & \textbf{73.44}         & \textbf{80.03}          & \textbf{73.52} & \textbf{80.02} & \textbf{73.53} & \textbf{80.00} & \textbf{73.54} \\
                                                & \nxk    & 79.75                   & 72.93                   & 79.87                  & 73.24                  & 79.92                   & 73.39          & 79.95          & 73.45          & 79.96          & 73.49          \\
                                                & \sxs    & 79.87                   & 73.17                   & 79.91                  & 73.29                  & 79.94                   & 73.37          & 79.96          & 73.44          & 79.97          & 73.49          \\
\midrule

\multirow{4}{*}{\rotatebox[origin=c]{90}{en$\to$ru}} & \ombr & 83.52                   & 77.01                   & 83.52                  & 77.01                  & 83.52                   & 77.02          & 83.52          & 77.02          & 83.52          & 77.01          \\
 \cmidrule{2-12}
                                                & PMBR   & \textbf{83.53}          & 75.94          & \textbf{83.63}         & \textbf{76.50}         & \textbf{83.63}          & \textbf{76.81} & \textbf{83.60} & \textbf{76.96} & \textbf{83.56} & \textbf{77.01} \\
                                                & \nxk    & 82.18                   & 74.96                   & 82.90                  & 75.99                  & 83.28                   & 76.59          & 83.45          & 76.84          & 83.48          & 76.93          \\
                                                & \sxs    & 83.39                   & \textbf{76.12}                   & 83.52                  & 76.43                  & 83.57                   & 76.63          & 83.59          & 76.79          & 83.57          & 76.91          \\
\midrule
\multirow{4}{*}{\rotatebox[origin=c]{90}{ru$\to$en}} & \ombr & 79.17                   & 75.57                   & 79.17                  & 75.57                  & 79.17                   & 75.57          & 79.17          & 75.57          & 79.17          & 75.57          \\
 \cmidrule{2-12}
                                                & PMBR   & \textbf{79.15}          & \textbf{75.15}          & \textbf{79.23}         & \textbf{75.40}         & \textbf{79.22}          & \textbf{75.48} & \textbf{79.19} & 75.51          & \textbf{79.18} & \textbf{75.56} \\
                                                & \nxk    & 78.71                   & 74.68                   & 78.99                  & 75.21                  & 79.08                   & 75.41          & 79.14          & \textbf{75.52} & 79.15          & 75.54          \\
                                                & \sxs    & 78.98                   & 75.01                   & 79.06                  & 75.20                  & 79.10                   & 75.34          & 79.13          & 75.43          & 79.15          & 75.52 \\

\bottomrule
\end{tabular}
\end{table}

\begin{table}
  \caption{Summary of the average scores of the full DeEn WMT 2022 pairs scored 1000 times using \metricx and \comet as evaluation metrics while varying the size of the hypothesis list}
  \label{table:all_ns_deen}

  \small
  \centering
  \setlength{\colskip}{7mm}
  \setlength{\tabcolsep}{4pt}
  \nprounddigits{3}
    \begin{tabular}{ll
      ll@{\hskip \colskip}
      ll@{\hskip \colskip}
      ll@{\hskip \colskip}
      ll@{\hskip \colskip}
      ll
    }
\toprule
                                                & budget & \multicolumn{2}{c@{\hskip \colskip}}{1/32} & \multicolumn{2}{c@{\hskip \colskip}}{1/16} & \multicolumn{2}{c@{\hskip \colskip}}{1/8} & \multicolumn{2}{c@{\hskip \colskip}}{1/4} & \multicolumn{2}{c}{1/2} \\

  \cmidrule(lr{\colskip}){3-4}
  \cmidrule(r{\colskip}){5-6}
  \cmidrule(r{\colskip}){7-8}
  \cmidrule(r{\colskip}){9-10}
  \cmidrule(r){11-12}

                                                &        & \multicolumn{1}{c}{C}                   & \multicolumn{1}{c@{\hskip \colskip}}{X}                 & \multicolumn{1}{c}{C}                  & \multicolumn{1}{c@{\hskip \colskip}}{X}                & \multicolumn{1}{c}{C}                   & \multicolumn{1}{c@{\hskip \colskip}}{X}        & \multicolumn{1}{c}{C}          & \multicolumn{1}{c@{\hskip \colskip}}{X}        & \multicolumn{1}{c}{C}          & \multicolumn{1}{c}{X}        \\
\midrule

\multirow{4}{*}{\rotatebox[origin=c]{90}{N=32}} &FMBR & NA              & NA              & 79.91          & 73.29          & 79.91          & 73.29          & 79.91          & 73.29          & 79.91          & 73.29  \\
\cmidrule{2-12}
&PMBR   & NA              & NA              & \textbf{79.87} & \textbf{73.01} & \textbf{79.97} & \textbf{73.26} & \textbf{79.99} & \textbf{73.36} & \textbf{79.95} & \textbf{73.34} \\
&NxK    & NA              & NA              & 79.46          & 72.27          & 79.74          & 72.84          & 79.84          & 73.09          & 79.88          & 73.22          \\
&SxS    & NA              & NA              & 79.61          & 72.56          & 79.73          & 72.82          & 79.82          & 73.04          & 79.87          & 73.17          \\
\midrule

\multirow{4}{*}{\rotatebox[origin=c]{90}{N=64}} &OFMBR & 79.97          & 73.45          & 79.97          & 73.45          & 79.97          & 73.45          & 79.96          & 73.45          & 79.97          & 73.45          \\
 \cmidrule{2-12}
&PMBR   & \textbf{79.90} & \textbf{73.07} & \textbf{79.98} & \textbf{73.30} & \textbf{80.02} & \textbf{73.44} & \textbf{80.02} & \textbf{73.47} & \textbf{80.00} & \textbf{73.47} \\
&NxK    & 79.43          & 72.25          & 79.76          & 72.91          & 79.88          & 73.20          & 79.93          & 73.35          & 79.96          & 73.40          \\
&SxS    & 79.73          & 72.80          & 79.82          & 73.04          & 79.88          & 73.19          & 79.92          & 73.30          & 79.95          & 73.38          \\

\midrule

\multirow{4}{*}{\rotatebox[origin=c]{90}{N=128}} &FMBR & 79.97          & 73.52          & 79.97          & 73.52          & 79.97          & 73.52          & 79.97          & 73.52          & 79.97          & 73.52          \\
 \cmidrule{2-12}
&PMBR   & \textbf{80.00} & \textbf{73.32} & \textbf{80.01} & \textbf{73.44} & \textbf{80.03} & \textbf{73.52} & \textbf{80.02} & \textbf{73.53} & \textbf{80.00} & \textbf{73.54} \\
&NxK    & 79.75          & 72.93          & 79.87          & 73.24          & 79.92          & 73.39          & 79.95          & 73.45          & 79.96          & 73.49          \\
&SxS    & 79.87          & 73.17          & 79.91          & 73.29          & 79.94          & 73.37          & 79.96          & 73.44          & 79.97          & 73.49          \\
\midrule
\multirow{4}{*}{\rotatebox[origin=c]{90}{N=256}} &FMBR & 79.96          & 73.60          & 79.96          & 73.60          & 79.96          & 73.60          & 79.96          & 73.60          & 79.96          & 73.60          \\
 \cmidrule{2-12}
&PMBR   & \textbf{80.02} & \textbf{73.44} & \textbf{80.03} & \textbf{73.53} & \textbf{80.02} & \textbf{73.55} & \textbf{80.01} & \textbf{73.58} & \textbf{80.00} & \textbf{73.61} \\
&NxK    & 79.86          & 73.28          & 79.90          & 73.42          & 79.94          & 73.49          & 79.95          & 73.52          & 79.96          & 73.56          \\
&SxS    & 79.94          & 73.36          & 79.96          & 73.45          & 79.97          & 73.49          & 79.98          & 73.54          & 79.97          & 73.56 \\
\bottomrule
\end{tabular}

\end{table}

\begin{table}
  \caption{Summary of the average scores of the full EnDe WMT 2022 with N=256 and r=1/16 pairs scored 1000 times using \metricx and \comet. The MQM scores are limited to 65 examples where all systems disagreed.}
  \label{table:mqm}
  \centering
  \begin{tabular}{llll}
\toprule
          & \comet  & \metricx & MQM           \\
          \toprule
FMBR & 83.33 & 77.15  & 1.169         \\
\midrule
PMBR      & 83.51 & \textbf{76.95}  & \textbf{1.370} \\
NxK       & 83.11 & 76.75  & 1.746          \\
SxS       & \textbf{83.59} & 76.79  & 1.566         \\
\bottomrule
\end{tabular}
\end{table}

\section{Results}
\label{sec:results}

The main experimental results are summarized in Table~\ref{table:all_lps} and Table~\ref{table:all_ns_deen}. In Table~\ref{table:all_lps}, we fix $N=128$ and we study the behaviour of each approximation method by limiting their budget to a fraction of the full computational cost on each language pair. The top row comprises the results obtained with the full MBR method (\emph{FMBR}) running on the complete list of $N$=$128$ candidates, and can be considered as an upper bound for the performance of each approximation method. The number of utility calls for FMBR is $128^2 = 16\, 384$. In tabe Table~\ref{table:all_ns_deen}, we fix the language pair to (de\lpfromto en) and we set $N$ to different values. This simulates the behavior of approximation methods as the candidate list grows. Similar results for (en\lpfromto de) are shown in Appendix~\ref{table:all_ns_ende}. MQM human evaluation results are summarized in Table~\ref{table:mqm}.

As measuring performance with the same metric we are optimizing for has the risk of overfitting, we mainly focus on \comet to assess translation quality. These are the main findings:

\paragraph{(1) PMBR outperforms all other tested approximation methods}
PMBR outperforms both the $N$x$K$ and $S$x$S$ approximation methods across language pairs, sample sizes and budgets. The gap between the approximation methods closes as the budget increases. Moreover the results in Table~\ref{table:all_ns} show that the same pattern holds when the size of the hypotheses list changes.

\paragraph{(2) PMBR is competitive to FMBR}
We can reduce the computational cost by up to $r=1/32$ with PMBR without any loss in translation quality as measured by \comet. 
Interestingly, we observe that \metricx scores slightly drop when reducing the budget. As this does not affect the final translation quality as measured by \comet, we argue that this is a good sign and PMBR acts as some kind of regularization.

\paragraph{(3) Human Evaluation confirms (1) and (2)}
To verify our findings based on \comet, we do run a MQM human evaluation with professional translators. Results are summarized in Table~\ref{table:mqm}. The results confirm our previous findings: (1) PMBR is the best approximation method when compared to $N$x$K$ and $S$x$S$, and (2) PMBR is getting close to the performance of FMBR.

\section{Conclusions}
\label{sec:conclusions}
In this paper we have shown the inherent low-rank structure of Minimum Bayes Risk (MBR) score matrices which we leveraged to develop an approximation method for MBR decoding that achieves competitive performance while significantly reducing computational complexity.
Our empirical results demonstrate the efficacy of this approach across diverse language pairs and evaluation metrics, suggesting its potential for wider application in machine translation and other natural language generation tasks.

Future research could explore the efficacy of alternative matrix completion algorithms to further enhance the low-rank approximation.
In addition, the observed low-rank property could be exploited to inform sampling strategies, potentially leading to more efficient and informative data collection for MBR decoding.
Another promising avenue is to investigate the applicability of this work to domains beyond natural language generation tasks.

\section{Limitations}
\label{sec:limitations}
While we have verified that the MBR matrices are low-rank, we did not conduct an empirical analysis on their coherence. A low-rank matrix is easier to complete if its energy spreads evenly across different coordinates. This property is captured by the notion of coherence \citep{candes2008exact}.

In this paper, we only run experiments with \metricx as utility function. The computational costs for computing all pairwise utility scores is expensive. However, we showed that the low-rank matrix structure holds for both \metricx and chrF which gives us confidence that PMBR will generalize regardless of the utility function.

Our human evaluation is limited in size because it is costly. With automatic metrics, we can simulate multiple runs of scoring the datasets but this is not feasible with human evaluations. Thus, we put less statistical significance on our human evaluation.

\newpage

\bibliographystyle{acl_natbib}
\bibliography{bibliography}

%%%%%%%%%%%%%%%%%%%%%%%%%%%%%%%%%%%%%%%%%%%%%%%%%%%%%%%%%%%%
\newpage

\appendix

\section{Appendix / supplemental material}

\subsection{DeEn graphs for all hypothesis size lists}

\begin{figure}[b]
  \centering
  \includegraphics[width=0.8\columnwidth]{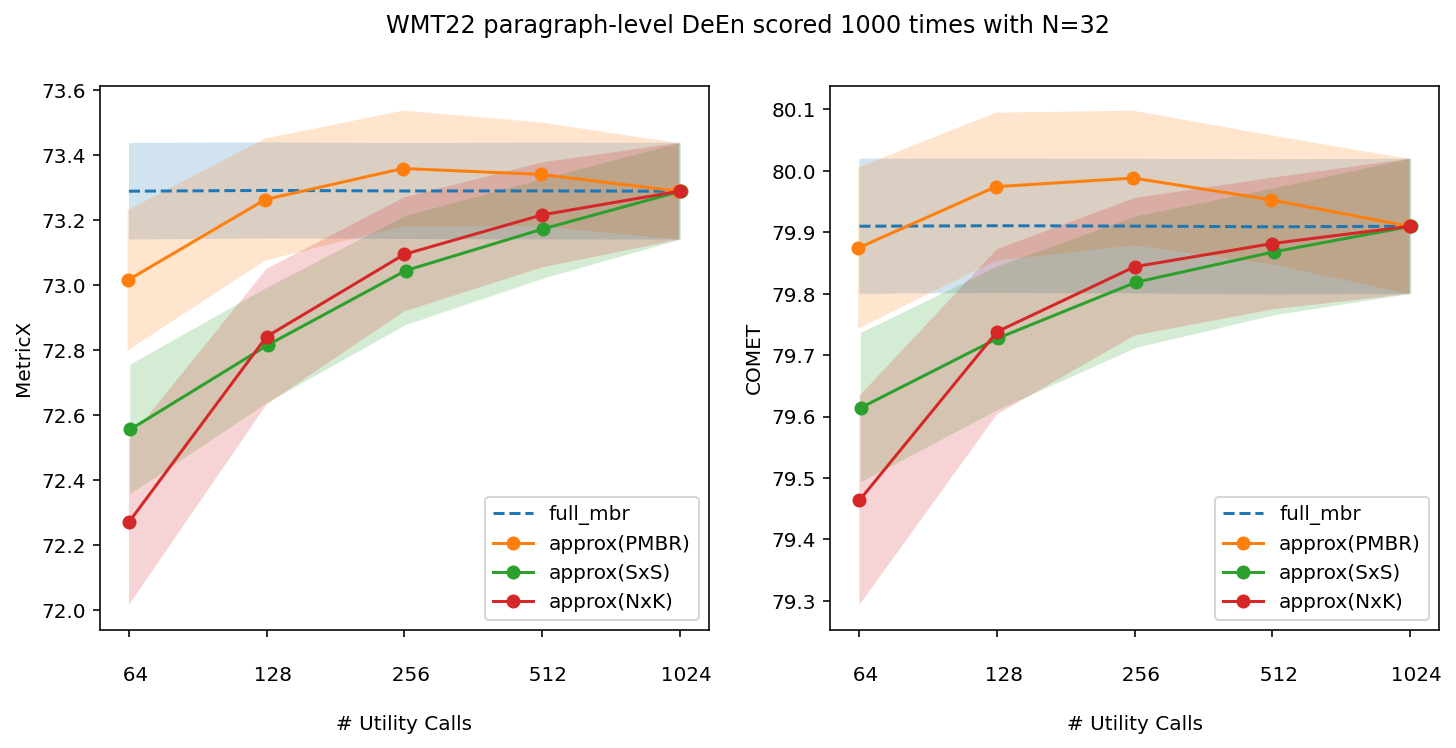}
 
  \caption{We scored WMT22 DeEn dataset 1000 times for each budget available. Each scoring picks without replacement 32 samples from the 1024 samples available for each sentence. The highlighted area shows the standard deviation of the scores.}
  \label{fig:deen_scores_32}
\end{figure}

\begin{figure}
  \centering
  \includegraphics[width=0.8\columnwidth]{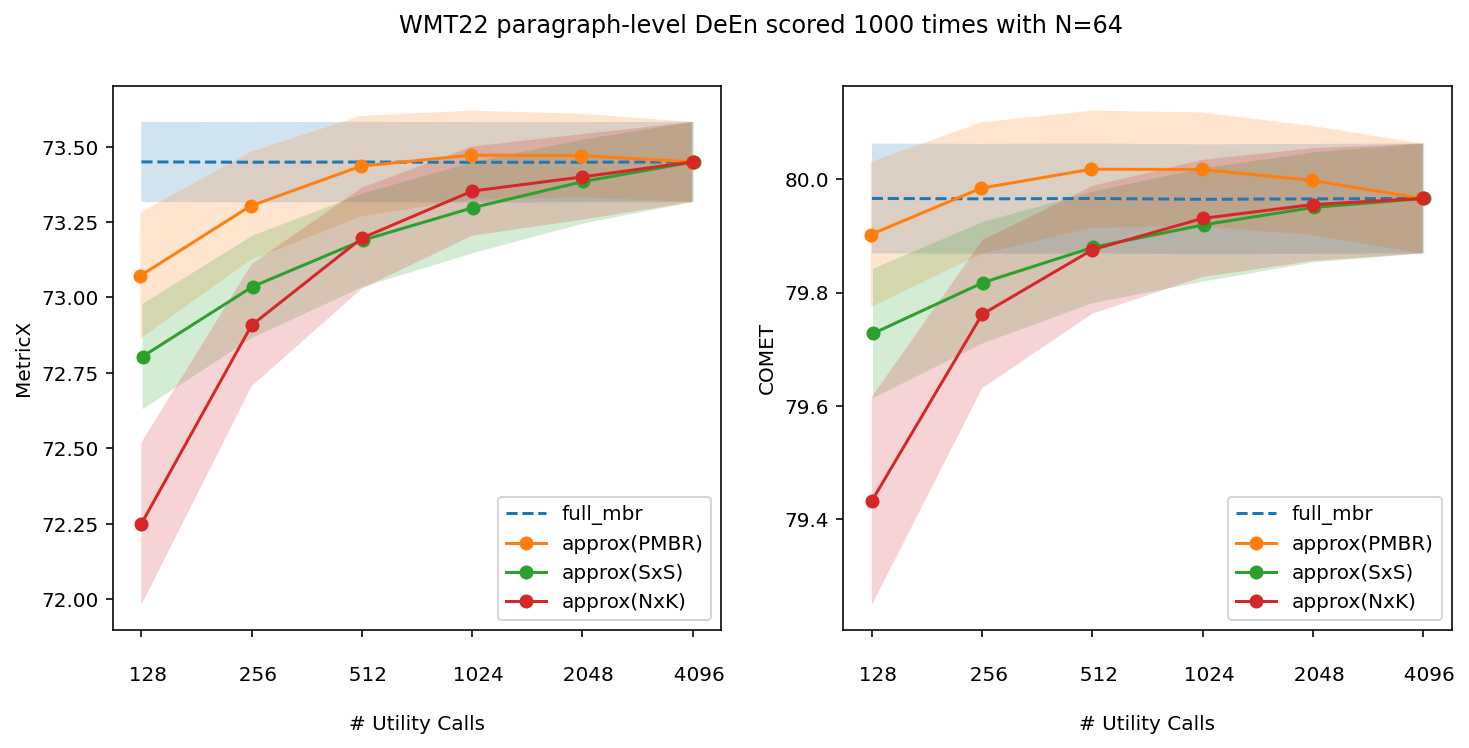}
 
  \caption{We scored WMT22 DeEn dataset 1000 times for each budget available. Each scoring picks without replacement 64 samples from the 1024 samples available for each sentence. The highlighted area shows the standard deviation of the scores.}
  \label{fig:deen_scores_64}
\end{figure}

\begin{figure}
  \centering
  \includegraphics[width=0.8\columnwidth]{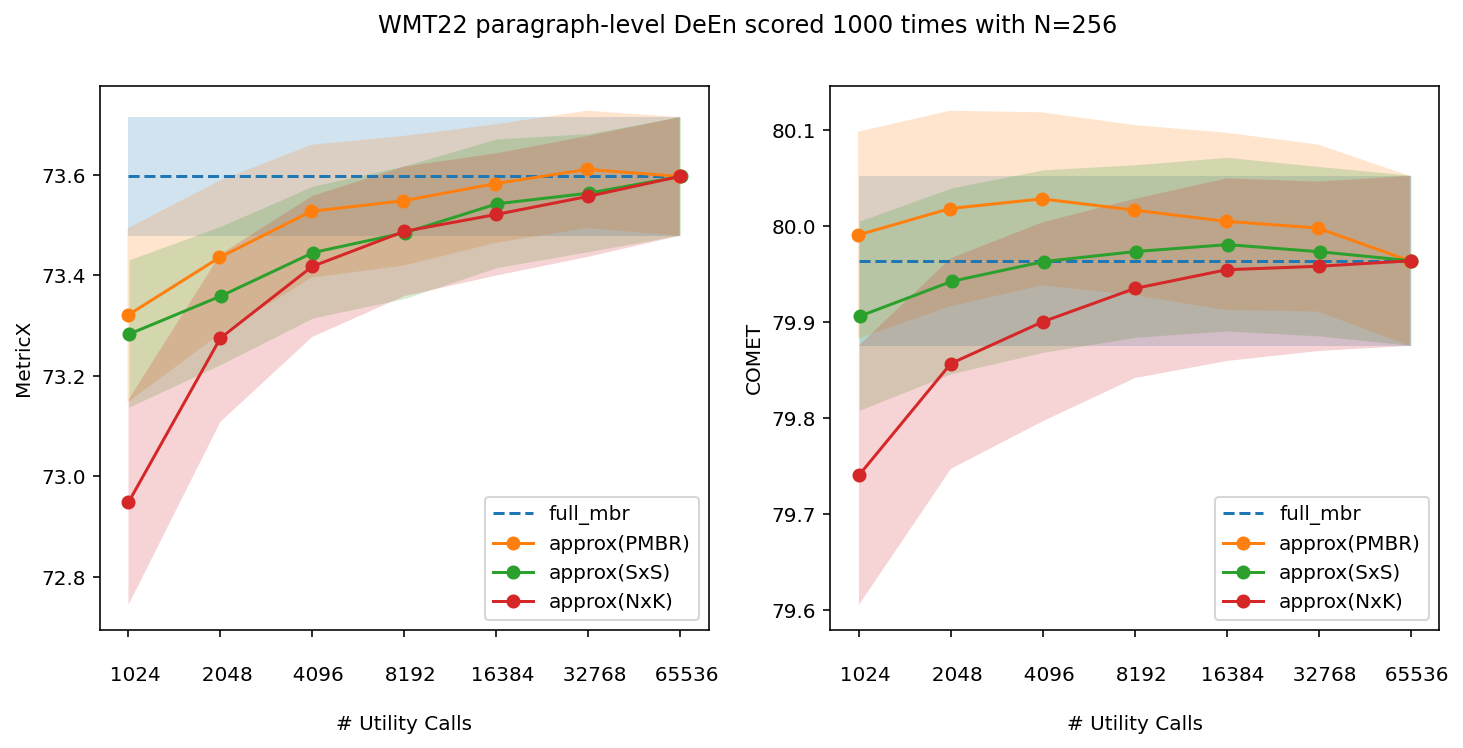}
 
  \caption{We scored WMT22 DeEn dataset 1000 times for each budget available. Each scoring picks without replacement 256 samples from the 1024 samples available for each sentence. The highlighted area shows the standard deviation of the scores.}
  \label{fig:deen_scores_256}
\end{figure}

Figures~\ref{fig:deen_scores_32}, \ref{fig:deen_scores_64} and \ref{fig:deen_scores_256} show the performance of PMBR for different hypothesis list sizes.

%%%%%%%%%%%%%%%%%%%%%%%%%%%%%%%%%%%%%%%%%%%%%%%%%%%%%%%%%%%%

% \usepackage{multirow}

\subsection{WMT 2022 paragraph level data statistics}
We provide the statistics of the dataset used after combining the sentences from the same document in table~\ref{table:testdata_stats}.

\begin{table}[b]
\label{table:testdata_stats}
\caption{Statistics of the WMT 2022 dataset and its paragraph level transformation.}
\centering
\begin{tabular}{llrrr}
\toprule
LP                      &                  & \# segments & Avg. \#words/segment & Avg. \#sent/segment \\
\toprule
                        & wmt\_22          & 2037        & 16.7            & 1.01           \\
\multirow{-2}{*}{En-De} & $\rightarrow$ paragraph & 219         & 154.6  & 7.99           \\
\midrule
                        & wmt\_22          & 2037        & 16.7            & 1.01           \\
\multirow{-2}{*}{En-Ru} & $\rightarrow$ paragraph & 219         & 154.6           & 7.99           \\
\midrule
                        & wmt\_22          & 1984        & 14.6            & 1.01           \\
\multirow{-2}{*}{De-En} & $\rightarrow$ paragraph & 309         & 93.2            & 5.73           \\
\midrule
                        & wmt\_22          & 2016        & 13.6            & 1.01           \\
\multirow{-2}{*}{Ru-En} & $\rightarrow$ paragraph & 258         & 106.0           & 7.32         \\
\bottomrule
\end{tabular}
\end{table}

\subsection{Scores for EnDE while varying the hypotheses list size}
In table~\ref{table:all_ns_ende}, we summarize all \comet and \metricx scores after varying the hypotheses list size.

\begin{table}
  \caption{Summary of the average scores of the full EnDe WMT 2022 pairs scored 1000 times using MetricX and COMET as evaluation metrics while varying the size of the hypothesis list}
  \label{table:all_ns_ende}
  \small
  \centering
  \setlength{\colskip}{7mm}
  \setlength{\tabcolsep}{4pt}
  \nprounddigits{3}
    \begin{tabular}{ll
      ll@{\hskip \colskip}
      ll@{\hskip \colskip}
      ll@{\hskip \colskip}
      ll@{\hskip \colskip}
      ll
    }
\toprule
                                                & budget & \multicolumn{2}{c@{\hskip \colskip}}{1/32} & \multicolumn{2}{c@{\hskip \colskip}}{1/16} & \multicolumn{2}{c@{\hskip \colskip}}{1/8} & \multicolumn{2}{c@{\hskip \colskip}}{1/4} & \multicolumn{2}{c}{1/2} \\

  \cmidrule(lr{\colskip}){3-4}
  \cmidrule(r{\colskip}){5-6}
  \cmidrule(r{\colskip}){7-8}
  \cmidrule(r{\colskip}){9-10}
  \cmidrule(r){11-12}

                                                &        & \multicolumn{1}{c}{C}                   & \multicolumn{1}{c@{\hskip \colskip}}{X}                 & \multicolumn{1}{c}{C}                  & \multicolumn{1}{c@{\hskip \colskip}}{X}                & \multicolumn{1}{c}{C}                   & \multicolumn{1}{c@{\hskip \colskip}}{X}        & \multicolumn{1}{c}{C}          & \multicolumn{1}{c@{\hskip \colskip}}{X}        & \multicolumn{1}{c}{C}          & \multicolumn{1}{c}{X}        \\
\midrule

\multirow{4}{*}{\rotatebox[origin=c]{90}{N=32}} &FMBR & NA              & NA              & 0.8352          & 0.7641          & 0.8352          & 0.7641          & 0.8352          & 0.7641          & 0.8352          & 0.7641          \\
\cmidrule{2-12}
&PMBR   & NA              & NA              & \textbf{0.8294} & \textbf{0.7489} & \textbf{0.8347} & \textbf{0.7571} & \textbf{0.8357} & \textbf{0.7615} & \textbf{0.8356} & \textbf{0.7634} \\
&NxK    & NA              & NA              & 0.8072          & 0.7262          & 0.8229          & 0.7445          & 0.8297          & 0.7541          & 0.8335          & 0.7595          \\
&SxS    & NA              & NA              & 0.8255          & 0.7463          & 0.8291          & 0.7525          & 0.8323          & 0.7577          & 0.8339          & 0.7611          \\
\midrule

\multirow{4}{*}{\rotatebox[origin=c]{90}{N=64}} &FMBR & 0.8361          & 0.7679          & 0.8360          & 0.7679          & 0.8360          & 0.7679          & 0.8361          & 0.7679          & 0.8359          & 0.7679          \\
 \cmidrule{2-12}
&PMBR    & \textbf{0.8304} & 0.7504          & \textbf{0.8354} & \textbf{0.7587} & \textbf{0.8366} & \textbf{0.7637} & \textbf{0.8366} & \textbf{0.7662} & \textbf{0.8365} & \textbf{0.7676} \\
&NxK    & 0.8057          & 0.7266          & 0.8226          & 0.7472          & 0.8299          & 0.7577          & 0.8338          & 0.7633          & 0.8354          & 0.7659          \\
&SxS    & 0.8293          & \textbf{0.7525} & 0.8323          & 0.7578          & 0.8340          & 0.7612          & 0.8352          & 0.7642          & 0.8356          & 0.7662          \\

\midrule

\multirow{4}{*}{\rotatebox[origin=c]{90}{N=128}} &FMBR & 0.8352          & 0.7701          & 0.8352          & 0.7701          & 0.8352          & 0.7702          & 0.8352          & 0.7702          & 0.8352          & 0.7701          \\
 \cmidrule{2-12}
&PMBR   & \textbf{0.8353} & 0.7594          & \textbf{0.8363} & \textbf{0.7650} & \textbf{0.8363} & \textbf{0.7681} & \textbf{0.8360} & \textbf{0.7696} & \textbf{0.8356} & \textbf{0.7701} \\
&NxK    & 0.8218          & 0.7496          & 0.8290          & 0.7599          & 0.8328          & 0.7659          & 0.8345          & 0.7684          & 0.8348          & 0.7693          \\
&SxS    & 0.8339          & \textbf{0.7612} & 0.8352          & 0.7643          & 0.8357          & 0.7663          & 0.8359          & 0.7679          & 0.8357          & 0.7691          \\
\midrule

\multirow{4}{*}{\rotatebox[origin=c]{90}{N=256}} &FMBR & 0.8332          & 0.7715          & 0.8333          & 0.7715          & 0.8332          & 0.7715          & 0.8332          & 0.7715          & 0.8332          & 0.7715          \\
 \cmidrule{2-12}
&PMBR   & \textbf{0.8354} & 0.7659          & \textbf{0.8351} & \textbf{0.7695} & \textbf{0.8347} & \textbf{0.7711} & \textbf{0.8341} & \textbf{0.7714} & \textbf{0.8338} & \textbf{0.7717} \\
&NxK     & 0.8274          & 0.7617          & 0.8311          & 0.7675          & 0.8326          & 0.7699          & 0.8332          & 0.7710          & 0.8332          & 0.7712          \\
&SxS    & 0.8357          & \textbf{0.7662} & 0.8359          & 0.7679          & 0.8357          & 0.7692          & 0.8351          & 0.7702          & 0.8344          & 0.7709  \\
\bottomrule

\end{tabular}

\end{table}

\subsection{Compute Resources}
We give a high level estimate of the resources to run the experiments:

\begin{itemize}
    \item Samples Generation: We used around 500 TPUv5 for 5 hours per language pair to generate the samples.
    \item MetricX pairwise computations: We used around 2000 TPUv4 for 24 hours per language pair to compute all the scores.
    \item Scoring simulations: These were run on CPUs in parallel on a cluster of 1000 machines. Each setting (budget, hypothesis length) takes around 15 minutes to run.
\end{itemize}

\end{document}